\documentclass[letterpaper]{article} % DO NOT CHANGE THIS
\usepackage{aaai2026}  % DO NOT CHANGE THIS
\usepackage{times}  % DO NOT CHANGE THIS
\usepackage{helvet}  % DO NOT CHANGE THIS
\usepackage{courier}  % DO NOT CHANGE THIS
\usepackage[hyphens]{url}  % DO NOT CHANGE THIS
\usepackage{graphicx} % DO NOT CHANGE THIS
\usepackage{natbib}  % DO NOT CHANGE THIS AND DO NOT ADD ANY OPTIONS TO IT
\usepackage{caption} % DO NOT CHANGE THIS AND DO NOT ADD ANY OPTIONS TO IT
\usepackage{algorithm}
\usepackage{algorithmic}
\usepackage{latexsym}
\usepackage{subcaption}
\usepackage{siunitx}
\usepackage{booktabs}
\usepackage{microtype}
\usepackage{natbib}
\usepackage{multirow}
\usepackage{amsmath}

\usepackage{longtable}
\usepackage{arydshln}
\usepackage{newfloat}
\usepackage{listings}
\DeclareCaptionStyle{ruled}{labelfont=normalfont,labelsep=colon,strut=off} % DO NOT CHANGE THIS
\floatstyle{ruled}
\newfloat{listing}{tb}{lst}{}
\floatname{listing}{Listing}
\title{Toward Real-World Chinese Psychological Support Dialogues: CPsDD Dataset and a Co-Evolving Multi-Agent System}
\author {
    Yuanchen Shi\textsuperscript{\rm 1},
    Longyin Zhang\textsuperscript{\rm 2},
    Fang Kong\textsuperscript{\rm 1}\textsuperscript{\rm *}
}
\affiliations {
    \textsuperscript{\rm 1}School of Computer Science and Technology, Soochow University, Suzhou, China\\
    \textsuperscript{\rm 2}Aural \& Language Intelligence, Institute for Infocomm Research, A*STAR, Singapore\\
    20227927002@stu.suda.edu.cn, zhang\_longyin@i2r.a-star.edu.sg, kongfang@suda.edu.cn
}

\usepackage{bibentry}
\begin{document}

\maketitle

\begin{abstract}
The growing need for psychological support due to increasing pressures has exposed the scarcity of relevant datasets, particularly in non-English languages. To address this, we propose a framework that leverages limited real-world data and expert knowledge to fine-tune two large language models: Dialog Generator and Dialog Modifier. The Generator creates large-scale psychological counseling dialogues based on predefined paths, which guide system response strategies and user interactions, forming the basis for effective support. The Modifier refines these dialogues to align with real-world data quality. Through both automated and manual review, we construct the \textbf{C}hinese \textbf{P}sychological \textbf{s}upport \textbf{D}ialogue \textbf{D}ataset (CPsDD), containing 68K dialogues across 13 groups, 16 psychological problems, 13 causes, and 12 support focuses. Additionally, we introduce the \textbf{C}omprehensive \textbf{A}gent \textbf{D}ialogue \textbf{S}upport \textbf{S}ystem (CADSS), where a Profiler analyzes user characteristics, a Summarizer condenses dialogue history, a Planner selects strategies, and a Supporter generates empathetic responses. The experimental results of the Strategy Prediction and Emotional Support Conversation (ESC) tasks demonstrate that CADSS achieves state-of-the-art performance on both CPsDD and ESConv datasets. 

\end{abstract}


\section{Introduction}
The increasing societal pressures in modern life have led to a growing demand for psychological support, especially through dialogue systems \citep{radez2021children, divya2022suicidality}. Training dialogue systems that can provide psychological support has great potential for applications such as customer service chatbots, emotional counseling, and social media interactions \citep{zhou2020design}. In recent years, the Emotional Support Conversation (ESC) task has emerged, aiming to assist users in alleviating psychological problems and resolving challenges related to their emotional distress \citep{heaney2008social, ijcai2022p600}. However, due to the privacy concerns in psychological counseling and the high cost of manually collecting dialogues, datasets for the ESC task are extremely scarce \citep{liu-etal-2021-towards}, especially in the Chinese field.

\begin{figure}[!t]
\begin{center}
  \includegraphics[width=0.8\linewidth]{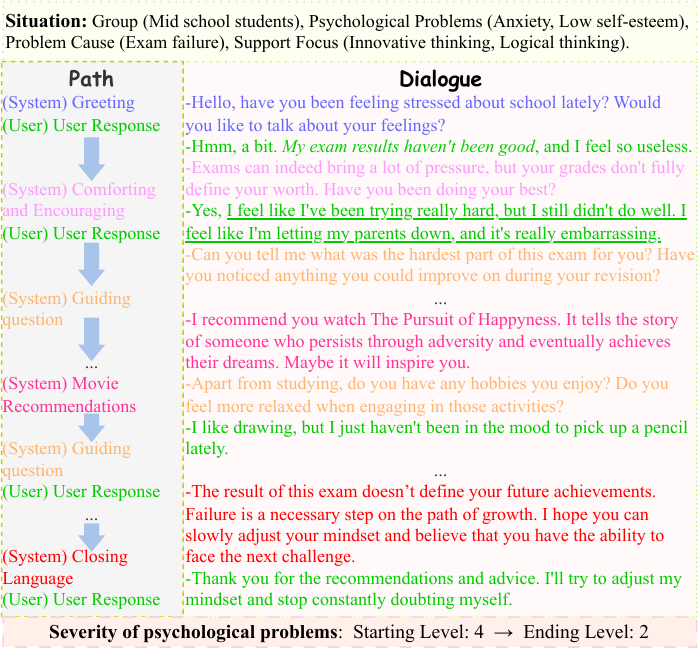} 
  \caption {An example from CPsDD dataset.}
  \label{F1}
\end{center}
\end{figure}

Large Language Models (LLMs) excel at generating data and providing empathetic responses \citep{zheng2023lmsys, sorin2024large}. However, in psychological counseling, LLMs often provide lengthy and formulaic replies, lacking empathy and substantial support \citep{wang-etal-2024-muffin}. To address the scarcity of Chinese psychological dialogue data and the challenge of generating real-world aligned data, we propose a framework combining real-world dialogues, expert knowledge, and LLM agents. Specifically, we collect expert-provided counseling dialogues and use GPT-4o with Chain-of-Thought (CoT) to generate similar data, which is then modified by psychological experts to meet real-world standards. Extensive research highlights the importance of appropriate strategies in providing psychological support \citep{cicognani2011coping, rachmad2022psychological, lutz2022prospective}. Therefore we employ experts to design 9 common response strategies, which we annotate for each system response. Strategies are extracted as dialogue paths, and along with user situations (see Figure \ref{F1}), which we use to fine-tune a Dialog Generator to guide dialogue generation. We also fine-tune a Dialog Modifier with expert-modified data to enhance data quality. Finally, we perform rigorous manual and LLM checks to ensure realism. Our framework significantly reduces resource and time consumption while ensuring high quality, compared to existing methods simulating user-system interactions \citep{he-etal-2024-planning, ye2024sweetiechat}.

We introduce the Chinese Psychological support Dialogue Dataset (CPsDD), which covers 13 groups, 16 psychological problems, 13 problem causes, and 12 support focuses, with a total of 68,136 dialogues. As shown in Figure \ref{F1}, each dialogue is annotated with its strategy path, user's situation, and changes in the severity of psychological problems before and after the counseling. We also propose the \textbf{C}omprehensive \textbf{A}gent \textbf{D}ialogue \textbf{S}upport \textbf{S}ystem (CADSS), which integrates four specialized agents: a Profiler that analyzes user characteristics, a Summarizer that condenses dialogue history, a Planner that selects strategies, and a Supporter that generates empathetic responses. Experiments show that CADSS outperforms existing models in both Strategy Prediction and ESC tasks on CPsDD and ESConv dataset.

Our main contributions are as follows:
\begin{itemize}
\item  We propose a framework that generates realistic, high-quality psychological dialogues by guiding dialogue paths and integrating expert knowledge, significantly reducing resource and manual consumption.
\item We introduce CPsDD, the first large-scale Chinese psychological support dialogue dataset for both strategy prediction and ESC tasks, which covers a wide range of common psychological groups, problems and causes.
\item We present CADSS, which achieves state-of-the-art results in both strategy prediction and ESC tasks on CPsDD and ESConv datasets. Our dataset and models will be publicly available.
\end{itemize}

\section{Related Work} 
\subsection{Psychological Support Dialogue Chatbot}
Open dialogues have long been a key method for addressing mental health issues \citep{razzaque2016introduction}, and with the rise of mobile devices, online psychological support has become more accessible \citep{pasikowska2013dialogue}. Early studies explored chatbots for psychiatric consultation, with \citet{oh2017chatbot} integrating natural language understanding and multimodal emotion recognition for emotional monitoring, and \citet{doi:10.1177/0706743719828977} investigating dialogue agents for diagnosing and treating mental illnesses like depression and anxiety.

As AI models advance, the feasibility of AI chatbots in psychological support has been further explored \citep{casu2024ai}. \citet{lee2024influence} proposed a text-based chatbot addressing privacy concerns, while \citet{chen2024psychatbot} developed PsyChatbot for depression, based on a retrieval-based QA algorithm within a cognitive behavioral therapy framework. Recent work, such as SoulChat \citep{chen-etal-2023-soulchat}, has enhanced LLMs' empathy and listening abilities, and EmoLLM \citep{liu2024emollms} addresses both affective classification and regression tasks to provide more comprehensive emotional support through improved sentiment analysis and emotion intensity detection.

\subsection{Psychological Dialogue Dataset} 
Psychological dialogue datasets are scarce due to the difficulty of data collection. \citet{rashkin2018towards} found that dialogue agents often struggle with emotion recognition, leading to the EmpatheticDialogues dataset with 25K emotion-contextual dialogues. \citet{welivita2021large} created a large-scale empathy response dataset, annotating 32 emotions and 8 empathy response intentions. \citet{zhang-etal-2024-escot} developed the ESCoT dataset with emotional support strategies enhanced by CoT, and \citet{liu-etal-2021-towards} introduced the ESConv dataset, rich in support strategy annotations. With the rise of LLMs, \citet{ye2024sweetiechat} used two LLMs to create the ServeForEmo dataset with 3.7K dialogues for the ESC task.

In the Chinese field, PsyQA \citep{sun-etal-2021-psyqa} focused on long counseling text for mental health support in a question-answering format. Recently, SmileChat \citep{qiu-etal-2024-smile} expanded single-turn dialogues into multi-turn conversations using ChatGPT, creating 55.2K dialogues for comprehensive mental health support. \citet{zhang-etal-2024-cpsycoun} introduced CPsyCounD, consisting of 3.1K dialogues, generated using a two-stage Memo2Demo approach where psychological supervisors first convert counseling reports into notes, followed by counselors generating multi-turn dialogues based on these notes.  \citet{li-etal-2023-understanding} developed an annotation framework for client reactions during online counseling, helping counselors adjust strategies. This paper proposes CPsDD, which builds on ESConv to fill the gap in Chinese psychological dialogue datasets with comprehensive strategy annotations.

\begin{figure*}[!t]
\begin{center}
  \includegraphics[width=0.7\linewidth]{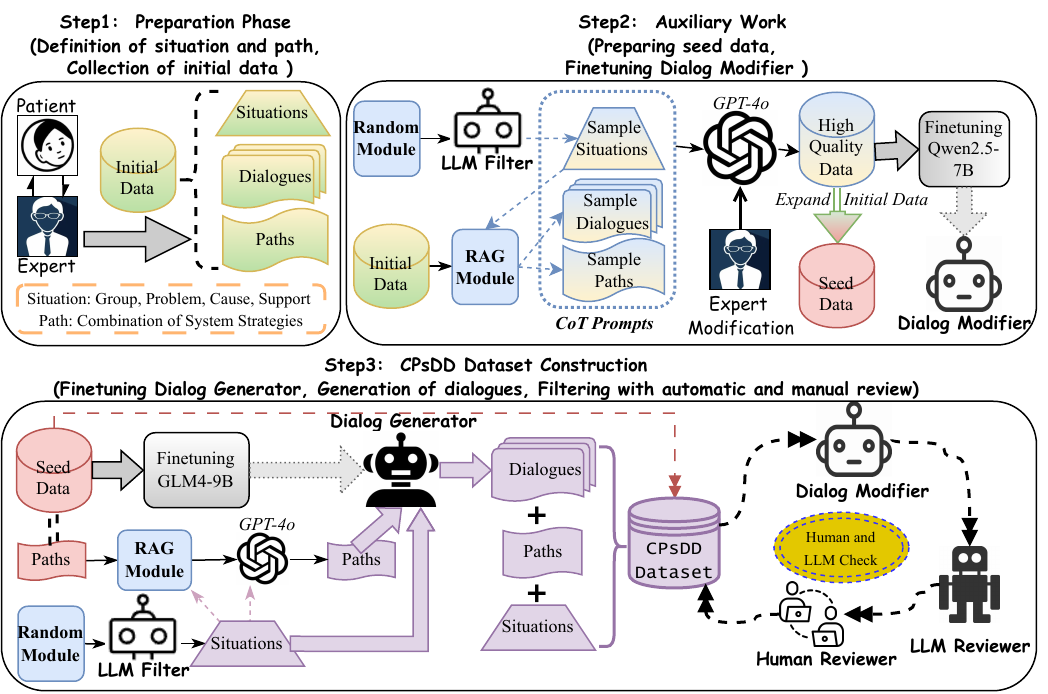} 
  \caption {The overall framework of constructing CPsDD dataset.}
  \label{constructCPsDD}
\end{center}
\end{figure*}

\section{CPsDD Dataset}
\subsection{Data Preparation and Preprocessing}

\begin{figure}[!t]
\begin{center}
  \includegraphics[width=0.8\linewidth]{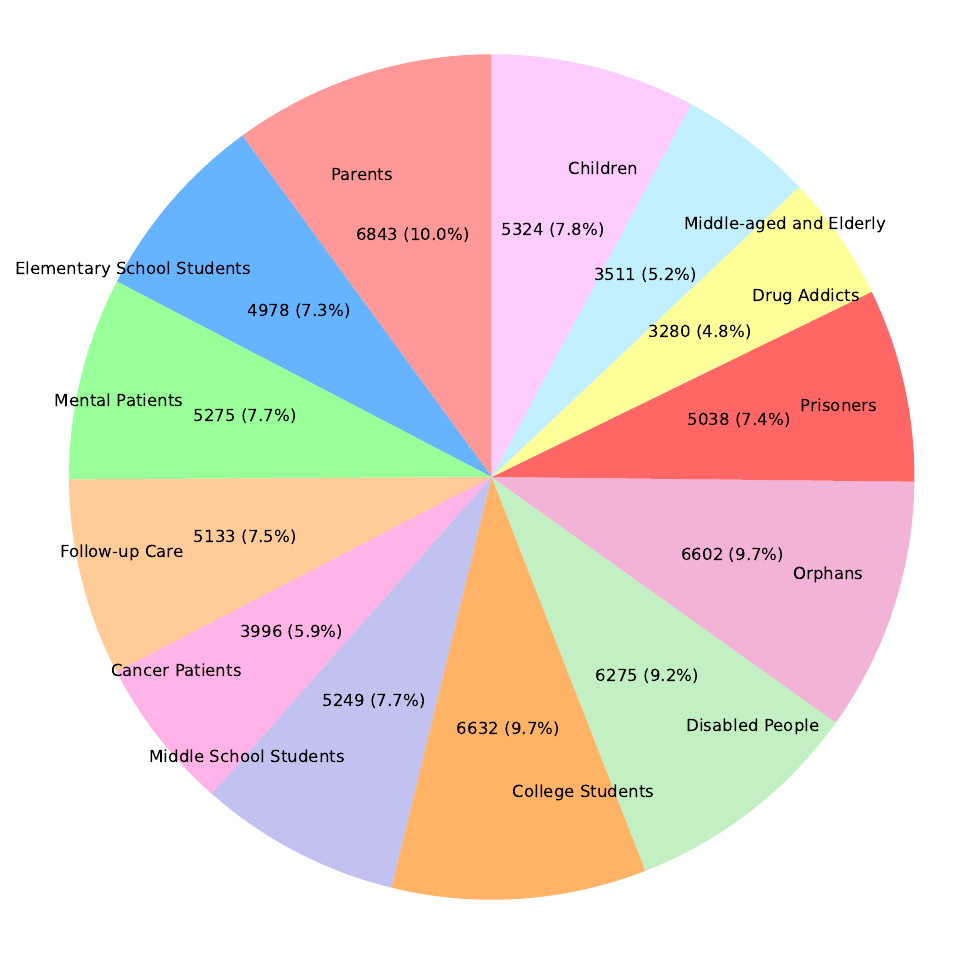} 
  \caption {Distribution of different groups in CPsDD.}
  \label{group}
\end{center}
\end{figure}

\begin{figure*}[!t]
    \centering
    \begin{subfigure}[b]{0.4\textwidth}
        \centering
        \includegraphics[width=\textwidth]{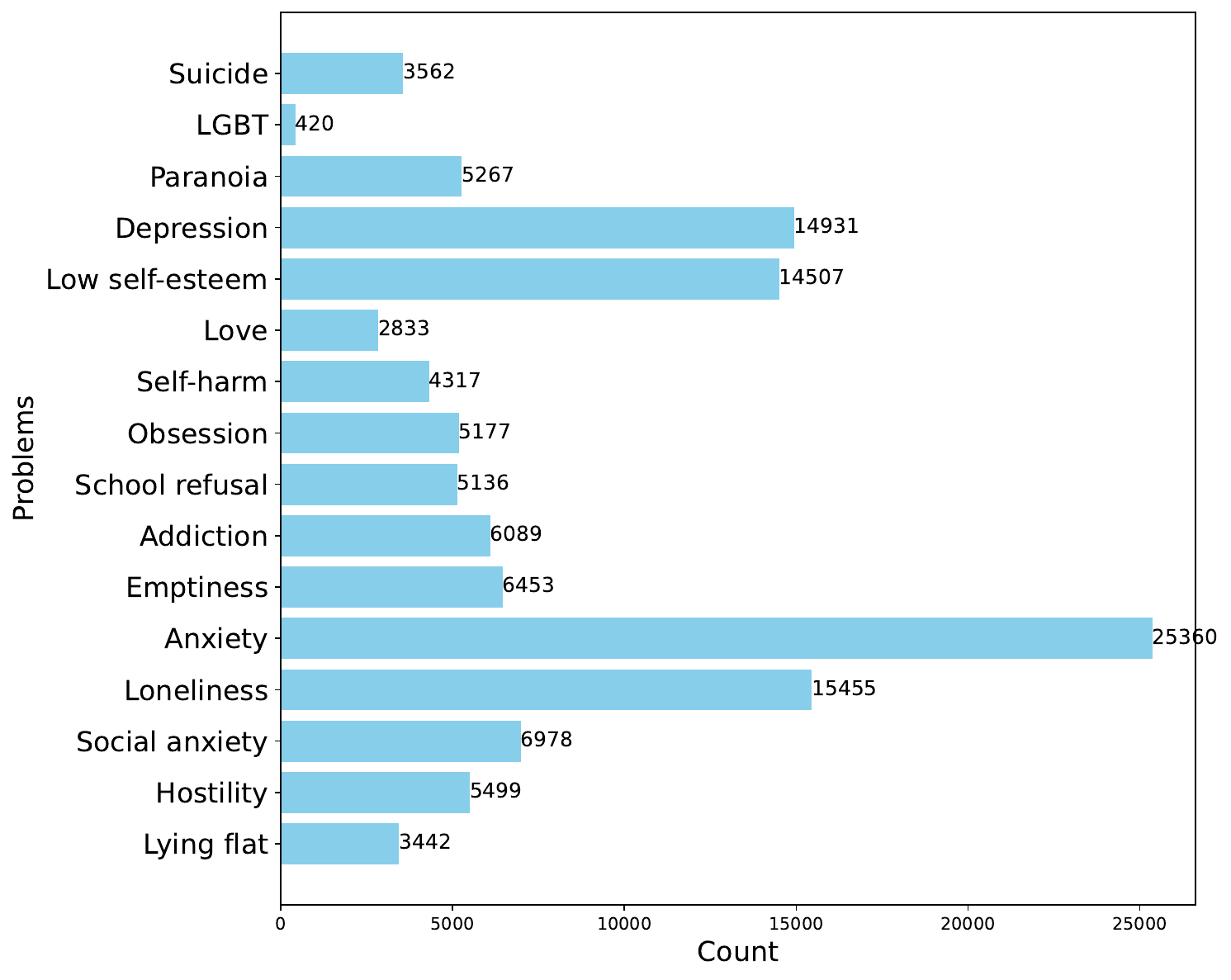}
        \caption{Frequency of different psychological problems.}
        \label{a}
    \end{subfigure}
    \hspace{0.05\textwidth} % 控制左右间距
    \begin{subfigure}[b]{0.4\textwidth}
        \centering
        \includegraphics[width=\textwidth]{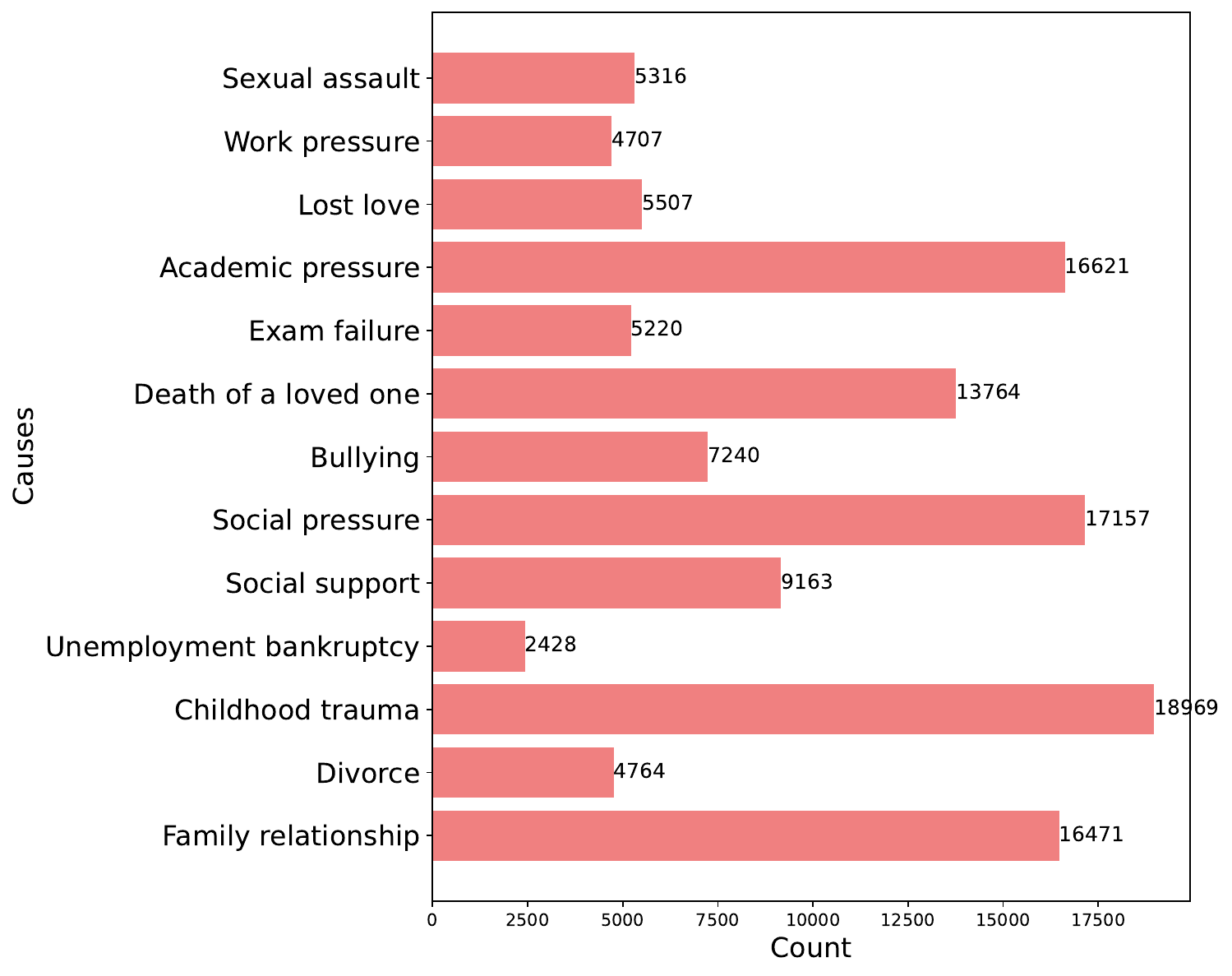}
        \caption{Frequency of different problem causes.}
        \label{b}
    \end{subfigure}
    
    \vskip\baselineskip % 控制上下间距

    \begin{subfigure}[b]{0.4\textwidth}
        \centering
        \includegraphics[width=\textwidth]{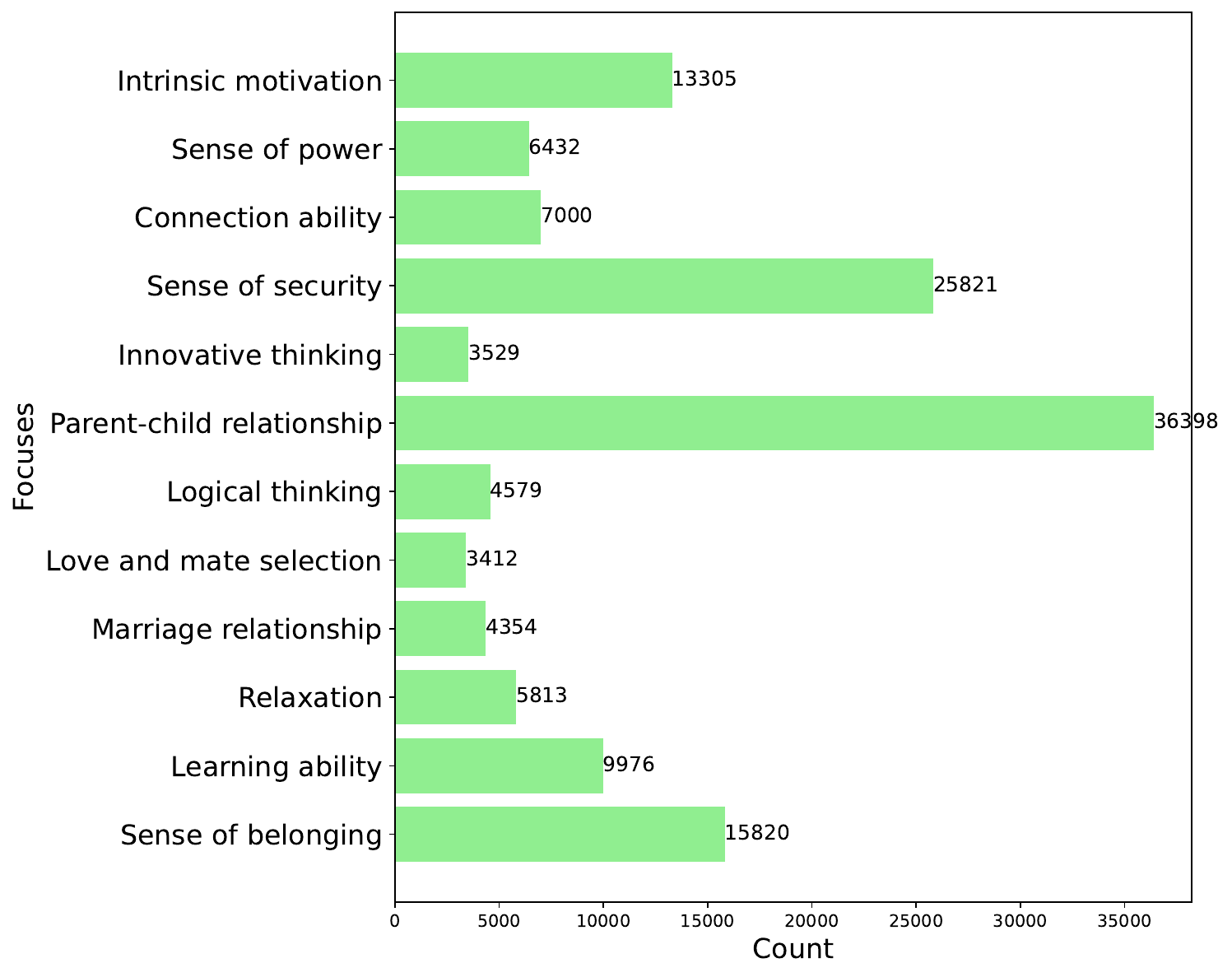}
        \caption{Frequency of psychological support focuses.}
        \label{c}
    \end{subfigure}
    \hspace{0.05\textwidth} % 控制左右间距
    \begin{subfigure}[b]{0.4\textwidth}
        \centering
        \includegraphics[width=\textwidth]{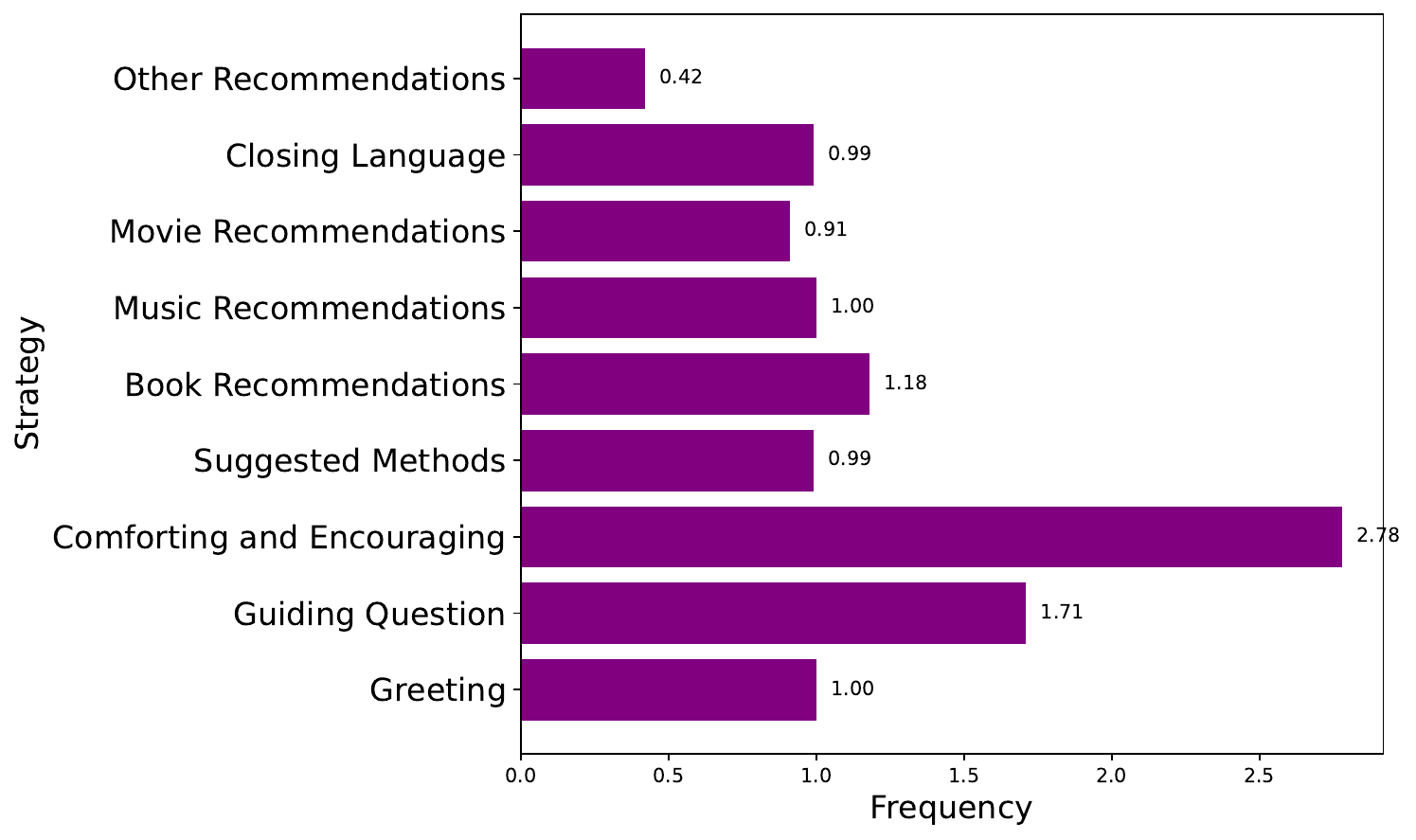}
        \caption{Average frequency of different response strategies.}
        \label{d}
    \end{subfigure}

    \caption{The frequency distribution of problems, causes, focuses, and strategies in CPsDD.}
    \label{all}
\end{figure*}

As shown in Figure \ref{constructCPsDD}, we begins with the identification of 13 psychological disorder groups, 16 common psychological problems, 13 causes, and 12 support focuses, developed in collaboration with three certified psychological experts \citep{shek2002family, xiao2013methodology, pan2023unpacking}. These experts, with extensive experience in psychological counseling across various settings (e.g., prisons, schools, and universities), ensured broad coverage of user populations. Figures~\ref{group} and \ref{all} provide detailed lists of these elements. We collected 130 real psychological counseling dialogues from experts, ensuring privacy by anonymizing all data and obtaining informed consent from the original users. Each expert annotates the dialogues, labeling psychological problems, causes, and focuses based on the defined taxonomy.

Following previous work \citep{liu-etal-2021-towards}, we defined 9 response strategies, including practical recommendations for providing actionable solutions to psychological challenges. These strategies guide the dialogue flow, offering steps for emotional relief, with each response forming a structured path tailored to address user distress.

After preparing the initial data, we use a Random Module to generate 200K situation labels by randomly selecting one user group and 1-3 problems, causes, and focuses. We employ models such as GLM4-9B \citep{glm2024chatglm}, Qwen2.5-7B \citep{yang2024qwen2}, and DeepSeek-R1 \citep{liu2024deepseek} to evaluate these situations and filter out implausible ones, resulting in 135K valid situations. See Fig.1 in Supplementary Materials (SM) for details. We then ensure comprehensive coverage by selecting 20 situations per group, retrieving corresponding dialogues as demonstration examples, and forming CoT Prompts to guide GPT-4o in generating similar dialogues (Fig.2 in SM).

To maintain data quality and alignment with domain expertise, the generated dialogues are reviewed and modified by experts. For example, school-related dialogues are reviewed by school psychologists, while those involving prisoners are handled by prison counselors. This targeted revision minimizes workload while ensuring context-appropriate data. The resulting dialogues, along with the initial data, form the Seed Data for large-scale generation. Finally, we fine-tune Qwen2.5-7B using Low-Rank Adaptation (LoRA) \citep{hu2021lora} to create the Dialog Modifier, a model that efficiently integrates expert knowledge to modify psychological dialogues (Fig.5 in SM).

\subsection{Dataset Construction}
We fine-tune GLM4-9B using the Seed Data (Fig.4 in SM) with LoRA, resulting in a Dialog Generator capable of generating high-quality psychological dialogues based on user situations and dialogue paths. We randomly generate Situations, retrieve sample data from the Seed Data, and construct a CoT Prompt (Fig.3 in SM) for GPT-4o to generate a dialogue path for each Situation. Using these inputs, we generate a total of 135K psychological dialogues.

To improve data quality, all generated dialogues are first refined by the Dialog Modifier. Then, GPT-4o serves as the LLM Reviewer, scoring dialogues based on their effectiveness and consistency with the Situation. Dialogues scoring 9-10 are retained, those scoring 7-8 are further refined using Reviewer feedback, and those scoring 6 or below are discarded. After this, 85K dialogues are re-evaluated. For those scoring 7-8, domain experts manually revise them based on their specialization, ensuring context-appropriate revisions (e.g., school-related dialogues reviewed by a school psychologist). This iterative process continues until all dialogues achieve a high score of 9-10. Finally, dialogues with fewer than 10 utterances are filtered out. The resulting dialogues, combined with the Seed Data, form the CPsDD dataset, consisting of 68K high-quality psychological dialogues.

\begin{table*}[!t]  \small
  \centering
  \begin{tabular}{l|cccccc}
    \toprule
  \textbf{Datasets} & \textbf{Size}&\textbf{Utts. (System/User)}&\textbf{\#Dia.len (System/User)}&\textbf{\#Utt.len (System/User)}&\textbf{St.s} & \textbf{Lan.} \\
    \midrule
ESConv & 1.3K  &  29.3K (14.6K/14.6K)  &  22.54 (11.27/11.27)  & 21.17 (19.90/22.45) &8& EN \\
ServeForEmo &  3.8K &  62.9K (30.7K/32.1K)  &  16.73 (8.18/8.55)  &17.97 (15.25/20.56)&8& EN   \\
SmileChat &  55.2K & 628.3K(318.2K/310.1K)  & 11.39(5.77/5.62)  & 81.29(56.90/106.32) &  - & ZH  \\
SoulChat &  258.4K & 3.0M(1.5M/1.5M)  & 11.79(5.87/5.92)  & 65.61(89.98/41.41)  &  - & ZH  \\
CPsyCounD & 3.1K  &  48.9K(24.4K/24.4K)   &  15.60(7.80/7.80)   & 44.84(56.02/33.67)    & -  &  ZH \\
PsyDTCorpus & 5K & 84.0K(42.0K/42.0K) & 19.49(9.75/9.75) & 45.01(55.72/34.30) & - & ZH \\
CPsDD(ours) & 68.1K  &  1.3M(0.7M/0.6M)  & 19.43(10.99/8.44)  &61.01(83.25/32.04)&9 &  ZH  \\
\bottomrule
  \end{tabular}
  \caption{Overall statistical comparison. \textit{Utts.} represents Utterances, \textit{\#} denotes the average, \textit{Dia.Len} represents the number of utterances in a dialogue, \textit{Lan.} represents Language, and \textit{St.s} represents the number of Strategies.
}
  \label{statistics}
\end{table*}

\subsection{Statistics Comparison}

Table~\ref{statistics} compares CPsDD with major psychological dialogue datasets in both English and Chinese. CPsDD surpasses English datasets like ESConv~\citep{liu-etal-2021-towards} and ServeForEmo~\citep{ye2024sweetiechat} in scale and utterances, offering richer, more comprehensive responses that reflect deeper emotional support. Compared to recent Chinese datasets such as SmileChat~\citep{qiu-etal-2024-smile}, SoulChat~\citep{chen-etal-2023-soulchat}, CPsyCounD~\citep{zhang-etal-2024-cpsycoun}, and PsyDTCorpus~\citep{xie2024psydtusingllmsconstruct}, CPsDD shows competitive scale and content richness, with more diverse and detailed system strategies. Additionally, CPsDD covers a broader range of user groups and psychological issues, making it the first Chinese dataset with comprehensive strategy annotations for the ESC task.

\begin{figure}[!t]
\begin{center}
  \includegraphics[width=0.8\linewidth]{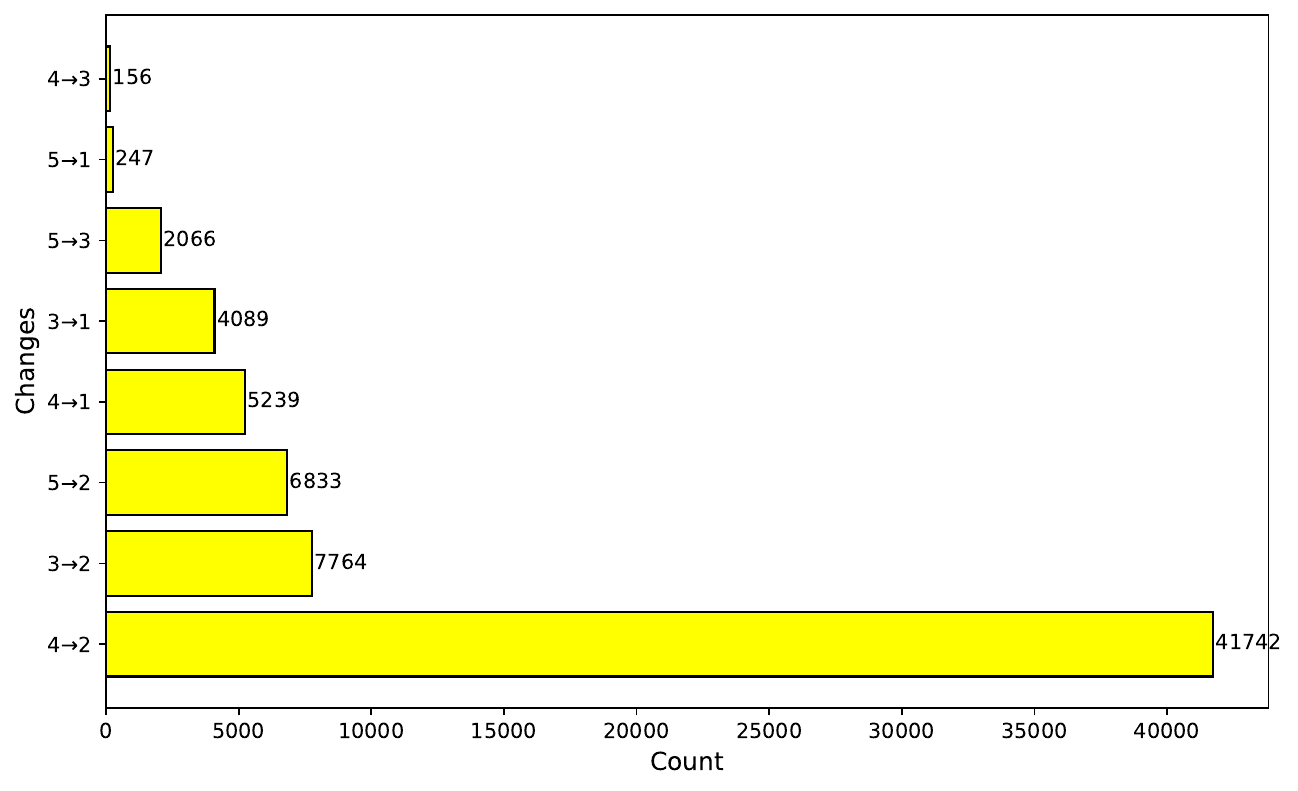} 
  \caption {Degrees of relief of psychological problems.}
  \label{Changes}
\end{center}
\end{figure}

\begin{figure}[!ht]
\begin{center}
  \includegraphics[width=0.8\linewidth]{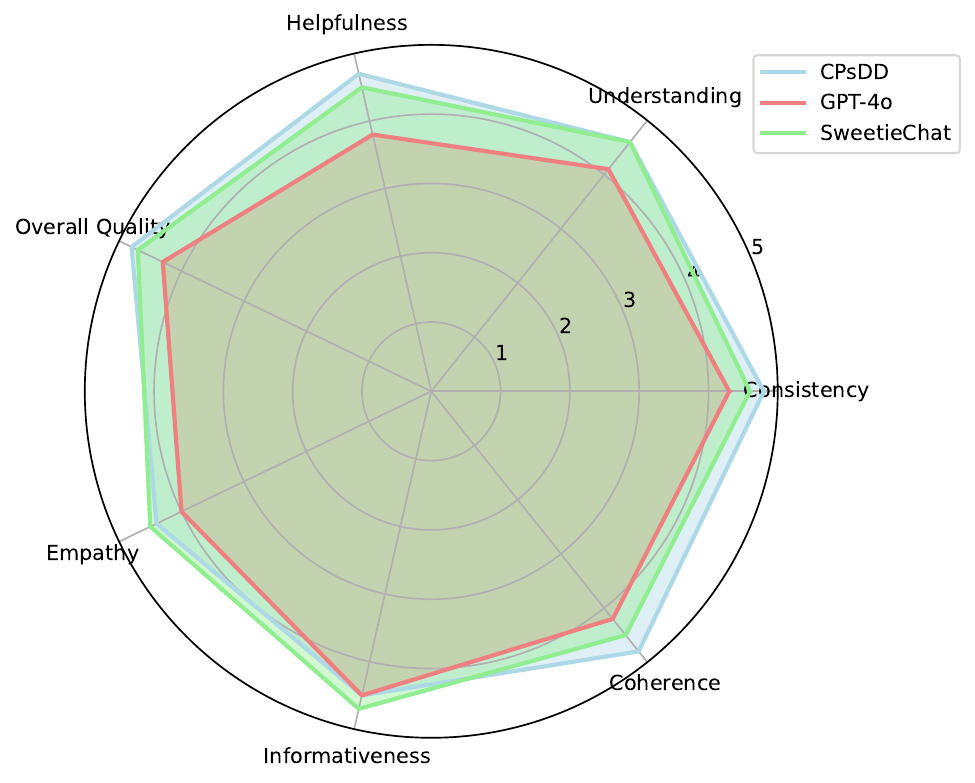} 
  \caption {Human evaluations of different methods for generating psychological dialogues.}
  \label{humanevaluate}
\end{center}
\end{figure}

\subsection{User Situation Distribution}

CPsDD includes 13 common groups. Figure \ref{group} shows a balanced distribution, with Parents most and Drug Addicts least represented. Figures \ref{a}, \ref{b}, and \ref{c} illustrate the frequency of various Problems, Causes, and Support Focuses. Anxiety, Loneliness, Depression, and Low Self-esteem are the most common problems, while Childhood Trauma, Social Pressure, Academic Pressure, and Family Issues are the primary causes. Psychological support mainly focuses on Parent-child relationships and Sense of security. These patterns reflect contemporary Chinese societal trends. Figure \ref{d} shows the average occurrence of each system strategy in dialogues, with Comforting and Encouraging being the most frequent, as they are central to providing psychological support.

\subsection{Psychological Support Effectiveness}
We use GPT-4o-mini as the Judger to assess the severity of the user's psychological problems based on the first three and last user responses in the dialogue (Fig.6 in SM). Figure \ref{Changes} shows that the majority of users' psychological issues are alleviated to a mild degree after the dialogue, with a few users being judged as having nearly recovered. Results show that most users' negative emotions are alleviated by at least two levels, further proving that the dialogues in CPsDD provide effective psychological support (Fig.7 in SM).

\subsection{Human Evaluation}
We randomly select 100 samples from CPsDD and used GPT-4o and SweetieChat \citep{ye2024sweetiechat} to generate 100 dialogues based on the situations. We employ 50 undergraduate students majoring in Chinese language to rate these dialogues on a scale of 1-5 in terms of Helpfulness, Understanding, Consistency, Coherence, Informativeness, Empathy, and Overall Quality (Tab.1 in SM). As shown in Figure \ref{humanevaluate}, CPsDD dialogues are more coherent, consistent with the situations, and more helpful. While SweetieChat outperforms in terms of informativeness and empathy, due to the nondeterminacy of ending interactions and the advantage of analyzing every responses, CPsDD receives the highest overall rating.

\section{Experiments}

\subsection{CADSS Framework}
\begin{figure}[!t]
\begin{center}
  \includegraphics[width=0.75\linewidth]{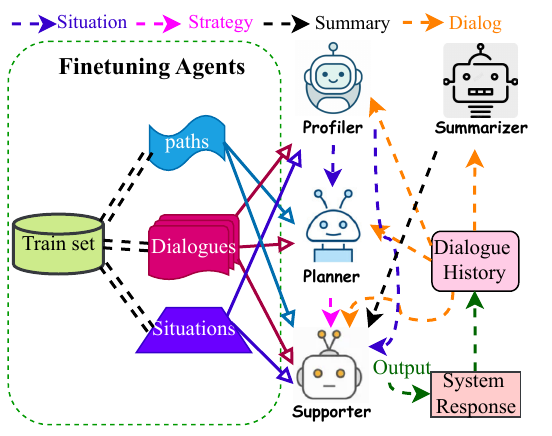} 
  \caption {The overall framework of CADSS.}
  \label{model}
\end{center}
\end{figure}

We propose the Comprehensive Agent Dialogue Support System (CADSS), as shown in Figure~\ref{model}. CADSS consists of four specialized agents: a \textbf{Profiler}, a \textbf{Summarizer}, a \textbf{Planner}, and a \textbf{Supporter}, all fine-tuned using Qwen2.5-7B models expect for the Summarizer which is used via prompting (Fig.8,9,10,11 in SM). The Profiler analyzes user characteristics and situations for downstream agents. The Summarizer condenses dialogue history and infers the user's emotion, intent, and psychological state. The Planner predicts the next response strategy based on the dialogue history and user situation. The Supporter then generates an empathetic response, guided by the selected strategy.

\begin{table*}[!t] \small
\centering
\begin{tabular}{l|c|cccccccccc}
\toprule
\textbf{Models} & \textbf{ACC$\uparrow$} & \textbf{PPL$\downarrow$} & \textbf{B-1$\uparrow$} & \textbf{B-2$\uparrow$} & \textbf{B-3$\uparrow$} & \textbf{B-4$\uparrow$} & \textbf{D-1$\uparrow$} & \textbf{D-2$\uparrow$}& \textbf{D-3$\uparrow$} & \textbf{R-L$\uparrow$}& \textbf{L-R} \\
\midrule
Qwen2.5& 26.01 &26.59 &20.27 &7.97 &4.51 &2.82 &81.83 &98.71 &99.75 &19.19 &1.12\\
DeepSeek-R1 &27.78 &-  & 18.92& 6.77& 3.51&1.74 &75.68 &97.96 & 99.85&18.52 &3.23\\
GPT-4o &33.17 & -& 21.26 &7.33 &3.44 &1.87 &74.31 &98.69 &99.89 &17.93 &1.83\\
CKPI\cite{hao-kong-2025-enhancing} & 47.93&22.76 &35.82  &26.47 &\textbf{20.62} &14.92 &48.48 & 87.53&95.35 &25.19 &0.63\\
SoulChat\cite{chen-etal-2023-soulchat}&-&25.93&20.86&7.82&4.00&2.33& 78.17&98.34& 99.68& 19.47& 1.10\\
MeChat\cite{qiu-etal-2024-smile}&-&23.59&19.86&7.42& 3.99& 2.43& 80.35& 98.60& 99.78& 19.25& 1.03\\
PsyChat\cite{qiu2024psychatclientcentricdialoguemental}&- &20.85 &18.29 &6.09 &2.86 &1.62 &76.20 &97.30 &99.10 &17.76 &1.16 \\
MindChat\footnote{https://github.com/X-D-Lab/MindChat}&- &\textbf{12.79} &20.32 &8.18 &4.53 &2.77 &74.42 &97.05 &99.14 &19.64 &1.43 \\
EmoLLM\cite{liu2024emollms} &- &14.09 &20.90 &9.71 &5.53 &3.31 &72.29 &97.82 &99.54 &21.48 &2.51 \\
CPsyCounX\cite{zhang-etal-2024-cpsycoun} &- &13.17 &20.59 &8.51 &4.57 &2.73 &78.54 &97.87 &99.49 &19.41 &1.34 \\
PsyDTLLM\cite{xie2024psydtusingllmsconstruct} &- &29.93 &19.62 &7.10 &3.51 &2.06 &\textbf{82.49} &98.91 &99.87 &19.14 &1.13  \\
CADSS(ours) &\textbf{80.98} &21.57 &\textbf{37.08} &\textbf{26.50} &20.47 &\textbf{16.54} &81.92 &98.92 &\textbf{99.90} &\textbf{37.97} &\textbf{1.00}  \\
\hline
w/o planner &- & 21.05&25.82 &13.80 &9.66 &7.25 &80.89 &99.05 &99.87 &25.71 &1.08 \\
w/o summarizer & 80.91&21.99 &36.22 &24.87 &19.63 &16.04 &81.46 &98.87 &99.83 &37.72 &0.96 \\
w/o profiler & 79.68&21.87 & 36.57&25.11 &20.06 &16.26 &81.43 &98.88 &99.84 &37.75 &0.99 \\
w/o planner\&summarizer &- &22.19 &25.40 &13.31 &9.24 &6.87 & 81.15&\textbf{99.07} &99.87 &25.26 &1.08 \\
w/o planner\&profiler & -&21.87 & 25.69&13.79 &9.67 &7.29 &80.54 &98.97 &99.84 &25.79 &1.12 \\
w/o summarizer\&profiler &79.75 &21.84 &36.16 &24.80 &19.56 &15.97 &81.24 &98.86 &99.83 &37.57 &0.97 \\
w/o all & - &23.77 & 25.19&13.10 &9.11 &6.79 & 80.91&99.04 &99.84 &24.93 &1.09\\
\bottomrule
\end{tabular}
\caption{Overall experimental results of different models on the Strategy Prediction and ESC tasks on CPsDD dataset. The upward and downward arrows represent which metric is better when higher or lower, respectively.}
\label{results}
\end{table*}

\begin{table*}[!t] \small
\centering
\begin{tabular}{l|c|cccccccccc}
\toprule
\textbf{Models} & \textbf{ACC$\uparrow$} & \textbf{PPL$\downarrow$} & \textbf{B-1$\uparrow$} & \textbf{B-2$\uparrow$} & \textbf{B-3$\uparrow$} & \textbf{B-4$\uparrow$} & \textbf{D-1$\uparrow$} & \textbf{D-2$\uparrow$} & \textbf{R-L$\uparrow$} \\
\midrule
BlenderBot-Joint\cite{liu-etal-2021-towards} & 17.69 & 17.39 & 18.78 & 7.02 & 3.20 & 1.63 & 2.96 & 17.87 & 14.92 \\
GLHG\cite{ijcai2022p600} & - & 15.67 & 19.66 & 7.57 & 3.74 & 2.13 & 3.50 & 21.61 & 16.37 \\
MISC\cite{tu-etal-2022-misc} & 31.67 & 16.27 & 16.31 & 7.31 & 3.26 & 2.20 & 4.62 & 20.17 & 17.51 \\
KEMI\cite{deng-etal-2023-knowledge} & - & 15.92 & - & 8.31 & - & 2.51 & - & - & 17.05 \\
TransESC\cite{zhao-etal-2023-transesc} & 34.71 & 15.82 & 17.92 & 7.64 & 4.01 & 2.43 & 4.73 & 20.48 & 17.51 \\
PAL\cite{cheng2023pal} & 34.51 & 15.92 & - & 8.75 & - & 2.66 & 5.00 & 30.27 & 18.06 \\
CKPI\cite{hao-kong-2025-enhancing} & 35.51 & \textbf{14.88} & 21.38 & 9.27 & 4.93 & 2.92 & 4.88 & 25.95 & \textbf{18.87} \\
Qwen2.5&14.11 & 40.60& 9.45&5.99 &4.32 &3.26 &7.22 &22.42 &9.27 \\
GPT-4o &23.72 &- & 15.42& 7.08&5.15 &3.87 &8.43 &29.61 &9.96 \\
DeepSeek-R1 &16.72 &- &10.96 &6.13 &3.86 &2.47 &6.69 &19.47 &7.13 \\

CADSS(ours) &\textbf{46.26} &30.14 &\textbf{27.54} &\textbf{15.06} &\textbf{10.05} &\textbf{5.64} &\textbf{8.88} &\textbf{32.49} &16.57 \\
\hline
w/o planner &- &31.81 &26.05 &13.47 &8.74 &4.61 &8.54 &31.87 &15.91 \\
w/o summarizer & 45.64&33.11 &26.99 &14.73&9.78 &5.40 &8.65 &31.28 &16.31 \\
w/o profiler &45.57 &35.51 & 27.17&14.80 &9.84 &5.45 &8.71 &31.81 &16.30 \\
w/o planner\&summarizer & -&38.65 &25.77 &13.51 &8.79 &4.62 &8.54 &30.00 &16.59 \\
w/o planner\&profiler &- &39.73 &25.72 &13.50 &8.78 &4.59 &8.54 &29.66 &16.51 \\
w/o summarizer\&profiler &45.53 &35.50 &27.07 &14.73 &9.76 &5.37 &8.66 &31.46 &16.29 \\
w/o all&- & 33.36&25.54 &13.25 &8.54 &5.41 &8.56 &30.05 &16.29 \\
\bottomrule
\end{tabular}
\caption{Overall experimental results on the Strategy Prediction and ESC tasks on the ESConv dataset.}
\label{ESConvresults}
\end{table*}

\subsection{Experimental Preparations}
\subsubsection{Data Process}
Experiments are conducted on both the CPsDD and ESConv datasets for the Strategy Prediction and Emotional Support Conversation (ESC) tasks. Each system response, along with its strategy, content, and dialogue history (excluding the initial greeting), is treated as a separate data instance. We combine both datasets, resulting in 0.7M instances which are split into train, dev, and test sets in an 8:1:1 ratio, to fine-tune the CADSS agents. This unified approach allows CADSS to leverage the strengths of both datasets, ensuring robust performance in multilingual and cross-cultural settings.

\subsubsection{Baseline Models}
For CPsDD dataset, we compare CADSS with several strong baselines, including many popular mental LLMs in Table \ref{results}. See Section.1 in SM for implementation details. For experiments on the ESConv dataset, we follow the latest SOTA setup in the CKPI benchmark, using the same set of strong baselines for comparison in Table \ref{ESConvresults}.

\subsubsection{Automatic Evaluation Metrics}
We use several automatic evaluation metrics: (ACC) measures accuracy in the strategy prediction task; (PPL) \citep{brown1992estimate} reflects the model's perplexity; BLEU-n (B-n) \citep{papineni2002bleu} and ROUGE-L (R-L) \citep{lin2004rouge} assess response quality; Distinct-n (D-n) \citep{li2015diversity} evaluates the proportion of unique n-grams; and Length-Ratio (L-R) \citep{tian2017make} calculates the ratio of response length to ground-truth length, helping determine whether the response is too concise or lengthy.

\begin{table*}[!t] \small
\centering
\begin{tabular}{|l|ccccccc|}
\toprule
\textbf{Models} & \textbf{Helpfulness}& \textbf{Understanding}& \textbf{Consistency}& \textbf{Coherence}& \textbf{Informativeness}&\textbf{Empathy}&  \textbf{Overall}\\
\midrule
GPT-4o&18 & 16&16 & 8&\textbf{25} &17 &17 \\
CKPI&7 &16 &11 & 13&9 &11 &2 \\
CADSS&\textbf{25} &\textbf{18} & \textbf{23}&\textbf{29} &16 &\textbf{22} & \textbf{31}\\
\bottomrule
\end{tabular}
\caption{Voting results of different model responses on each evaluation standard.}
\label{voting}
\end{table*}

\begin{figure*}[!t]
\begin{center}
  \includegraphics[width=0.7\linewidth]{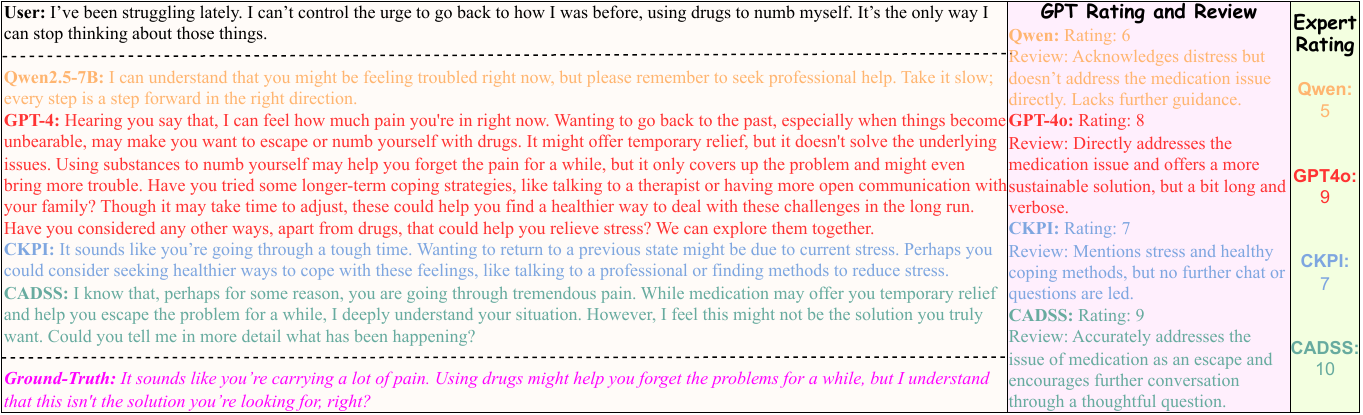} 
  \caption {Case study on different responses to user's psychological counseling, along with GPT and expert ratings.}
  \label{case}
\end{center}
\end{figure*}

\subsection{Experimental Results}
\subsubsection{CPsDD Results}
As shown in Table~\ref{results}, CADSS achieves state-of-the-art results on both the Strategy Prediction and ESC tasks on the CPsDD dataset, outperforming all baselines. Although CADSS's PPL is medium, it leads in strategy prediction accuracy, as well as in BLEU and ROUGE-L scores, reflecting its superior strategy control and generation quality. For diversity, CADSS, along with other LLMs like MeChat and PsyDTLLM, shows high Distinct-n scores, while CKPI lags behind due to its tendency to generate shorter or more repetitive responses. In terms of response length, CADSS's L-R score is close to 1, showing that its responses are well-aligned with the ground truth, whereas DeepSeek and CKPI generate overly long and shorter responses, respectively. Most mental LLMs, which lack strategy annotations, perform poorly in controllability and interpretability, further highlighting the importance of strategy integration for effective psychological support. Overall, these results demonstrate the effectiveness of CADSS's multi-agent design in producing rich, diverse, and user-aligned psychological support in real-world Chinese counseling scenarios.

\subsubsection{ESConv Results}
As shown in Table \ref{ESConvresults}, CADSS also achieves SOTA performance on the Strategy Prediction task of ESConv. In the ESC task, CADSS has a higher PPL score, as the LLM may exhibit more diversity and creativity when generating text, leading to a higher perplexity. CADSS significantly outperforms existing models in BLEU scores, demonstrating strong syntactic fluency and semantic accuracy. However, in the ROUGE-L score, CADSS slightly lags behind, as the higher diversity and creativity might result in a lower longest common subsequence (LCS) similarity with the reference text. In the Distinct score, CADSS also achieves the best performance, benefiting from the strong generation ability of the LLM backbone.

\subsubsection{Ablation Study}
We conduct ablation studies on both CPsDD and ESConv to evaluate the contribution of each agent in CADSS. As shown in Tables~\ref{results} and~\ref{ESConvresults}, removing any single agent leads to clear drops in strategy prediction accuracy and response quality (BLEU, ROUGE-L). The Planner is especially critical: without it, strategy prediction and generation metrics decrease most sharply, and responses become longer and less focused. The Profiler and Summarizer also play important roles in maintaining personalization and coherence. Notably, Distinct scores remain stable, reflecting the inherent diversity of LLM-based models, while higher PPL in ablated models indicates less confident and less controlled generation. These results confirm that all agents are indispensable for achieving strong and robust performance on both Chinese and English psychological support tasks.

\subsubsection{Manual Voting Results}
We organize 50 Chinese college students, each providing a psychological counseling utterance to different models for a response. After receiving the responses, they voted for the model they considered best based on various metrics (Tab.2 in SM). The results in Table \ref{voting} show that CADSS received the most votes in most metrics, particularly in Overall Quality, indicating it provides more comprehensive psychological support. While some response strategies limited the diversity of the responses, leading to slightly lower recognition in Informativeness, CADSS overall proves to be a more professional psychological support system that better meets user needs.

\section{Case Study}
Based on the same user input, we generate psychological support responses using different models, as shown in Figure \ref{case}. The original Chinese utterances available in Fig.12 in SM. Both GPT-4o and human experts rate the responses. CADSS receives the highest scores, as it fully understands the user's situation and provides reasonable psychological support, while also guiding the user to better express themselves. GPT-4o performs well, offering solutions and emotional value, but its responses are overly detailed, which could reduce the user's engagement. CKPI understands the user but lacks follow-up questions to drive the conversation forward. Qwen2.5-7B offers fairly standard responses with limited psychological support. We also deploy CADSS on a webpage, providing a ChatGPT-like interactive experience for psychological support. For further details and ethical considerations, refer to Section.2 in SM.

\section{Conclusion}

To address the increasing prevalence of psychological problems and the scarcity of related datasets, we introduce a novel framework that integrates expert knowledge with LLM reasoning through a path-guided dialogue generation approach. We present CPsDD, the first large-scale Chinese psychological support dialogue dataset with fine-grained strategy annotations, supporting both Strategy Prediction and ESC tasks. To further advance the field, we propose CADSS, a multi-agent system that achieves state-of-the-art results in generating high-quality psychological support responses across both Chinese and English datasets. In the future, CADSS can be further scaled with multimodal data.


\bibliography{AnonymousSubmission/LaTeX/acl}

\end{document}